\def\year{2021}\relax

\documentclass[letterpaper]{article} 
\usepackage{arxiv}  
\usepackage{times}  
\usepackage{helvet} 
\usepackage{courier}  
\usepackage[hyphens]{url}  
\usepackage{graphicx} 
\usepackage{natbib}  
\usepackage{caption} 

\usepackage{algorithm2e}
\usepackage{amsmath}
\usepackage[capitalize]{cleveref}
\usepackage{multirow}
\usepackage{booktabs}
\usepackage{mathtools}

\usepackage[super]{nth}
\DeclareMathOperator{\BERT}{BERT}
\DeclareMathOperator{\MLP}{MLP}
\DeclareMathOperator{\Relu}{Relu}

\usepackage{algorithm2e}
\usepackage{xcolor}
\usepackage{soul}
\usepackage{footmisc}

\newcommand{\xvar}[1]{\textsf{#1}}

\newcommand{\xvbox}[2]{\makebox[#1][l]{#2}}

\SetAlFnt{\footnotesize}

\SetAlCapSty{xAlCapSty}

\SetCommentSty{xCommentSty}

\SetNlSty{mynlfont}{}{} 

\LinesNumbered

\SetSideCommentRight

\DontPrintSemicolon

\RestyleAlgo{algoruled}

\usepackage{algorithm2e}

\usepackage{microtype}

\title{Investigating Gender Bias in BERT}

\author{
    Rishabh Bhardwaj,
    Navonil Majumder,
    Soujanya Poria
    \\
}

\affiliations{
    DeCLaRe Lab\\
    Singapore University of Technology and Design\\
    Singapore
}

\begin{document}

\maketitle

\begin{abstract}
Contextual language models (CLMs) have pushed the NLP benchmarks to a new height. It has become a new norm to utilize CLM provided word embeddings in downstream tasks such as text classification. However, unless addressed, CLMs are prone to learn intrinsic gender-bias in the dataset. As a result, predictions of downstream NLP models can vary noticeably by varying gender words, such as replacing ``he" to ``she", or even gender-neutral words. In this paper, we focus our analysis on a popular CLM, i.e., $\BERT$. We analyse the gender-bias it induces in five downstream tasks related to emotion and sentiment intensity prediction. For each task, we train a simple regressor utilizing $\BERT$'s word embeddings. We then evaluate the gender-bias in regressors using an equity evaluation corpus. Ideally and from the specific design, the models should discard gender informative features from the input. However, the results show a significant dependence of the system's predictions on gender-particular words and phrases. We claim that such biases can be reduced by removing gender-specific features from word embedding. Hence, for each layer in BERT, we identify directions that primarily encode gender information. The space formed by such directions is referred to as the gender subspace in the semantic space of word embeddings. We propose an algorithm that finds fine-grained gender directions, i.e., one primary direction for each BERT layer. This obviates the need of realizing gender subspace in multiple dimensions and prevents other crucial information from being omitted. Experiments show that removing embedding components in such directions achieves great success in reducing BERT-induced bias in the downstream tasks.

\end{abstract}

\section{Introduction}
\noindent Gender stereotypes can obstruct gender neutrality in many areas such as education, work, politics. Despite years of headway towards gender neutrality, the significant bias in social norms still exists. Automatic machine learning systems are likely to reproduce and reinforce existing gender stereotypes. Such issues have percolated down to even the language models that have recently set the state of the art in various natural language processing (NLP) tasks. However, the blunt application of language models risks introducing gender-bias in real-world systems.

It is becoming increasingly common to use an LM's contextualized word-vectors in downstream tasks such as text classification, question-answering, and conference resolution. In this work, we focus our analysis on one of the most famous language models: Bidirectional Encoder Representations for Transformers known as $\BERT$~\cite{devlin2018bert}. $\BERT$ is a transformer-based architecture \cite{vaswani2017attention} that has inspired many recent advances in machine learning even beyond language-only systems \cite{lu2019vilbert}. $\BERT$ allows parallelized training and deals with long-range dependencies better than RNN-based models such as ELMo.

Existing studies mostly focus on identifying gender-bias in context-independent word representations such as GloVe~\cite{bolukbasi2016man}. Contrarily, BERT word to vector(s) mapping is highly context-dependent which makes it difficult to analyse biases intrinsic to $\BERT$. We hypothesize that such biases will be reflected in downstream tasks exploiting BERT word embeddings. Hence, in this work, we investigate gender-bias induced by BERT in 5 downstream tasks that collectively fall in a category of tasks--Affect in tweets. The category splits into two sub-categories 1) emotion intensity and 2) valence (sentiment) intensity regression.

To perform the above-mentioned tasks, we train simple MLP-regressors exploiting BERT embeddings. We probe gender-bias in the trained models using an equity-evaluation corpus. The corpus consists of sentences especially designed to tease-out biases in NLP systems. Ideally, MLPs should not base their predictions on gender-specific words or phrases in the input. However, we observe the MLPs to consistently assign higher (or lower) scores to the sentences with words or phrases indicating a particular gender. For instance, one of the MLP regressors predicts high emotion intensity scores to sentences with female words than male words under the same context~\cite{poria2020beneath}. We call such systems as \textbf{gender-biased}. It is worth noting that the gender inclination is found to be specific to the task and word embedding used, hence ungeneralizable.

Due to the simplicity of MLP's, we hold BERT accountable for the observed gender-bias. Subsequently, we show the existence of layer-specific orthogonal directions where BERT encodes crucial gender information. We call such direction as \textbf{gender directions} and the space spanned by them as \textbf{gender subspace}. The directions (thus subspace) is unique to a BERT layer. The quality of extracted gender directions is identified by defining a new metric \textbf{gender separability}. To reduce the number of dimensions of gender subspace, we propose a novel algorithm that identifies fine-grained gender directions, i.e., one for every $\BERT$ layer. Thus, the obtained gender subspaces are 1-dimensional. The layer-wise elimination of vector components in gender directions helps reduce gender-bias in the downstream regression models.

To establish the importance of extracted gender directions, we design another downstream task, i.e., gender classification. The task specifically needs gender encoded features from the input word's vector representation. We find the BERT-based $\MLP$ to outperform a baseline gender classifier, proving the existence of gender-rich features in BERT embeddings. Additionally, removing gender-specific directional components from BERT embeddings drops the classification performance significantly. This concludes that the identified directions are close to the directions in BERT embeddings that encode the notion of gender.

\section{Related Work}
While a lot has been studied, identified, and mitigated when it comes to gender-bias in static word embeddings \cite{bolukbasi2016man, zhao-etal-2018-learning, caliskan2017semantics, zhao-etal-2018-gender}, very few recent works studied gender-bias in contextualized settings. We adapt the intuition of possible gender subspace in $\BERT$ from \cite{bolukbasi2016man}, which studied the existence of gender directions in static word embeddings. \cite{zhao-etal-2019-gender, basta-etal-2019-evaluating, gonen-goldberg-2019-lipstick} focused their study on ELMo. \cite{kurita-etal-2019-measuring} provided a template-based approach to quantify bias in BERT. \cite{sahlgren-olsson-2019-gender} studied bias in both contextualized and non-contextualized Swedish embeddings.

To the best of our knowledge, we are the first to identify gender-bias in BERT by analysing its impact on downstream tasks. We propose a novel algorithm to identify fine-grained gender directions to minimize the exclusion of important semantic information. Empirically, the elimination of embedding components in gender directions proves to be significantly reducing gender-bias in the tasks under study.

\section{Background}
\paragraph{BERT} In our study, we analyse $\BERT$ base–12 layers (transformer blocks), 12 attention heads, and 110 million parameters. The model is pre-trained on masked-language model and next sentence prediction tasks \cite{devlin2018bert} on lower-cased English text. Out-of-vocabulary (OOV) words are WordPiece tokenized that breaks a word into subwords from the pre-defined vocabulary. An input sequence of words $W$ is prepended with \texttt{[CLS]}, appended with \texttt{[SEP]}, and tokenized to generate $W_t = \{w_{cls}, w_1, ..., w_n, w_{sep}\}$. First, tokens are mapped to context-independent vectors $W_0=\{t_0^{cls}, t_0^0, \ldots, t_0^n, t^{sep}_0\}$, we denote it as $layer_0$. $\{layer_i\}_{i=1}^{12}$ are transformer layers that map vectors in $W_0$ to contextualized vectors $\{W_i\}_{i=1}^{12}$. We denote $t_j^i \in {\rm I\!R}^{d_b}$ as vector representation of $w_j$ at the output of $layer_i$. We utilize vocabulary and $\BERT$ pre-trained model from \cite{Wolf2019HuggingFacesTS}.

\section{Equity Evaluation}

\paragraph{Equity Evaluation Corpus (EEC)} ~\cite{kiritchenko-mohammad-2018-examining} 
The dataset contains template-based sentences such as ``$<$Name$>$ feels angry". $<$Name$>$ can be a female name such as ``Jasmine", or a male name such as ``Alan". An NLP-model is then asked to predict the intensity of emotion - angry. A system is called gender-biased when it consistently predicts higher/lower scores for sentences carrying female-names than male-names, or vice versa. The EEC contains 7 templates of type: $<$person$>$ and $<$emotion$>$. The place of variable $<$person$>$ can be filled by any of 60 gender-specific names or phrases. Out of 60, 40 are gender-specific names (20-female, 20-male). Rest 20 are noun phrases, particularly, 10 female-male pairs such as ``my mother" and ``my father". Variable $<$emotion$>$ can replace four emotions--Anger, Fear, Sadness, and Joy--each having 5 representative words~\footnote{eg:- \{angry, enraged\} represents a common emotion, i.e., anger}.
Thus, we have 1200 ($60 \times 5 \times 4$) samples for each template. In total we have 8400 ($7 \times 1200$) samples equally divided in female and male-specific sentences ($60 \times(5 \times 4)\times 7=4200$ each) and 4-emotion categories ($5 \times 7 \times 60=2100$ each). We refer readers to ~\shortcite{kiritchenko-mohammad-2018-examining} for an elaborate explanation.

\paragraph{Evaluation Methodology}
To evaluate an NLP system for its intrinsic gender-bias, we follow the same evaluation scheme as in \cite{kiritchenko-mohammad-2018-examining}. For a given template T and emotion word E, i.e., T-E format in EEC, we obtain 11 pairs of female-male intensity scores. One of the pairs is obtained by averaging the system's intensity predictions of input sentences with gender-specific names. The score pair consists of an average female score as its first element and a male score as its second. The other 10 scores are calculated from 10 noun phrase pairs used in the same T-E format. Thus for 7 templates and 20 emotion words, we have $7{\times}{20}{\times}11$ = 1540 pairs of scores. We define $\Delta_{F{\uparrow}{-}M{\downarrow}}$ as the average difference in female to male scores for those pairs with higher predicted intensity for females; vice versa to this defines $\Delta_{M{\uparrow}{-}F{\downarrow}}$. The number of occurrences where female scores are higher ($\#_{F{\uparrow}{-}M{\downarrow}}$), male scores are higher ($\#_{M{\uparrow}{-}F{\downarrow}}$), and both female-male scores are equal ($\#_{F{=}M}$) are also kept for a fine-grained evaluation.

\section{BERT Induced Bias} \label{EvalEEC_biased}
As mentioned earlier, we hypothesize that downstream tasks are prone to acquire gender-bias from $\BERT$ word embeddings. However, it is possible that a task-specific model enhances or diminishes the $\BERT$ induced bias, or learns its own bias. To prevent such scenarios, for all the tasks, we use shallow MLP regressors without fine-tuning $\BERT$ parameters. The simplicity of regressors will expose inherent gender-bias in $\BERT$. Next, we elaborate on the downstream tasks, the $\MLP$ architecture, and the training procedure.

\subsection{Downstream Tasks} 
Our bias evaluations are based on regression subtasks of SemEval-2018 Task 1--Affect in Tweets~\cite{mohammad-etal-2018-semeval}. The sub-tasks are majorly classified in:
\begin{itemize}
 
 \item[1] Emotion intensity regression ($\text{E}_\text{R}$-tasks): Given a tweet and an affective dimension $\in$ \{joy, fear, sadness, anger\}, determine emotion intensity I--a real-valued score between 0 (low mental state) and 1 (high mental state).
 
 \item[2] Sentiment intensity regression ($\text{S}_\text{R}$-task): Similar to $\text{E}_\text{R}$ tasks, given a tweet, determine intensity I of the sentiment.
\end{itemize}

For all the five tasks, i.e., 4-$\text{E}_\text{R}$ tasks and an $\text{S}_\text{R}$ task, train and test sets are provided with gold intensity scores.

\begin{figure}
    \centering
    \includegraphics[width=0.25\textwidth]{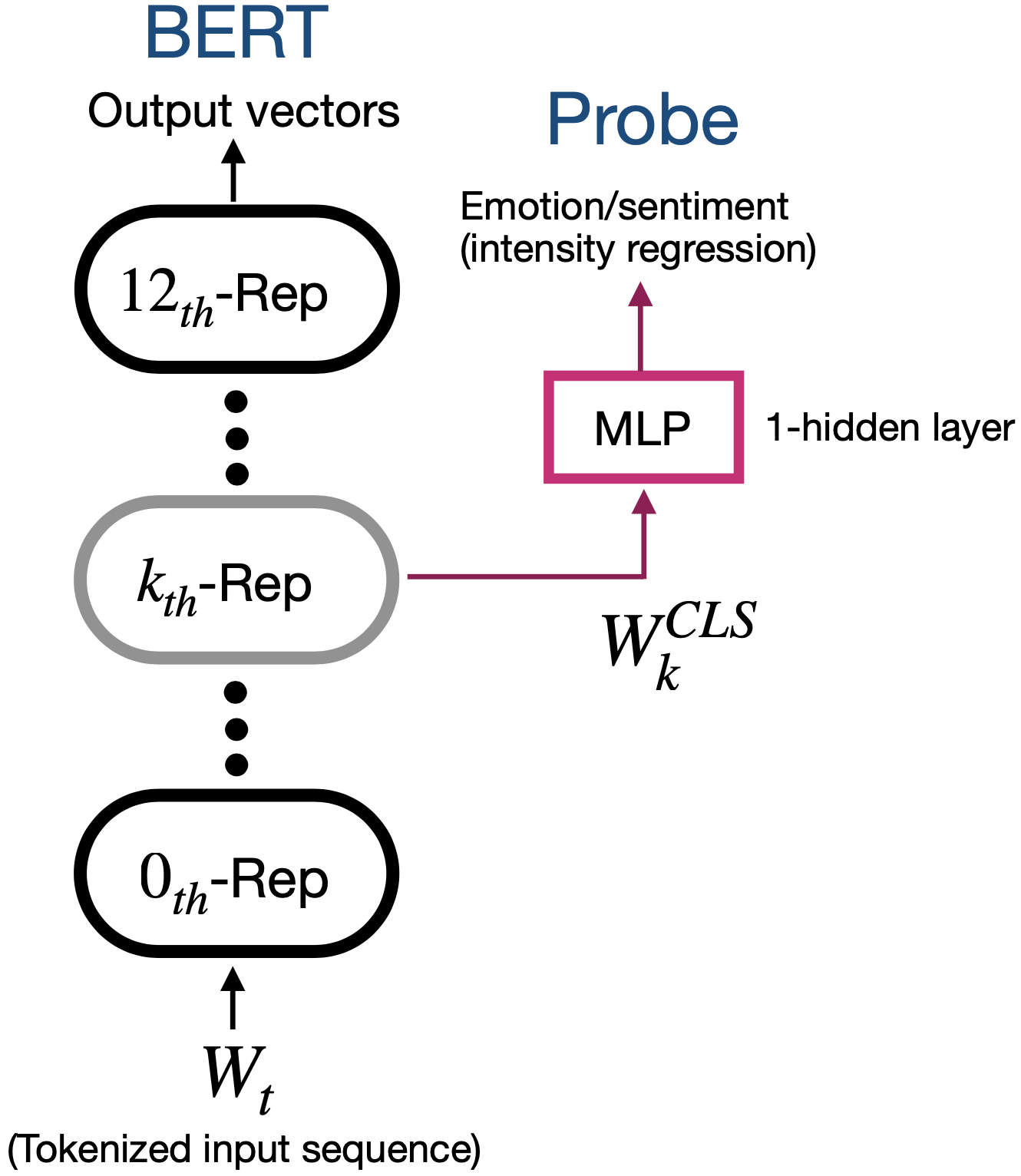}
    \caption{\small{MLP regressor trained on five individual regression tasks. Input to the MLP is 768-dimensional vector mapping of $\texttt{[CLS]}$ token at $\BERT$ $layer_k$.}}
    \label{fig:mlp_regressor}
\end{figure}

\subsection{BERT-based MLP regressor (BERT-MLP)} \label{BERT-MLP}
Since the \texttt{[CLS]} token was specifically introduced as a representative of the input sequence, it is reasonable to use its vector representations in downstream tasks (shown in \cref{fig:mlp_regressor}). For the sequence $W$, we obtain a deep contextualized vector representation of \texttt{[CLS]}, i.e., $W_k^{CLS}$ from $layer_k$ \footnote{$W_k^{CLS}$ and $t_k^{cls}$ represent the same vector.}.

For each task, we train a 2-layer $\MLP$ regressor to predict intensity $I_k \in [0,1]$ expressed by the sequence $W$, i.e., a tweet. Hence we train-test 5 different regressors. Input to the $\MLP$ is $W_k^{CLS} \in {\rm I\!R}^{768}$, hidden layer carries 200 neurons fully connected to the input vector, followed by $\Relu$ activation. The output is just an affine combination of the values obtained after activation. The Adam optimizer minimizes squared-loss between the $\MLP$ outputs and ground-truth intensity values. We divide the dataset into batches of 200 samples. In each iteration, parameters are updated to reduce loss accumulated in one batch. We score the models by calculating Pearson's correlations between predicted and expected intensity values \footnote{Following~\url{https://competitions.codalab.org/competitions/17751#results}}. The $\MLP$ architecture is kept simple to decipher features encoded in $\BERT$ word representations. Thus, the task performance of an $\MLP$ will rely heavily on the features provided by $\BERT$. $\BERT$ columns in \cref{tab:EEC_Biased_Emt} and \ref{tab:EEC_Biased_Emt_pref} show Pearson scores. We restrict our analysis to embeddings from deep layers in $\BERT$ i.e., $layer_{11}$ and $layer_{12}$.

\subsection{Equity Evaluation of BERT-MLP}
We evaluate $layer_{11}$ and $layer_{12}$ embedding of $\BERT$ separately. As shown in the \cref{tab:EEC_Biased_Emt} and \ref{tab:EEC_Biased_Emt_pref}, in columns correspond to $\BERT$, all five regressors show significant $\Delta$ values \footnote{Delta values can be compared to models studied in \shortcite{kiritchenko-mohammad-2018-examining}}. For each task, we observe that the regressors consistently assign high values to either of the genders. Moreover, not many cases are seen where $\MLP$'s assign equal scores to both the genders, i.e., $\#_{F{=}M}$. We discuss the results in later sections.

\section{Gender Debiasing} \label{DebBert}
In this section, we aim to uncover principal directions where $\BERT$ layers encode gender information. We hypothesize that removing the word vector components from gender directions will lead to reduced gender-bias in downstream tasks utilizing $\BERT$ embeddings.

\begin{itemize}
 \item[1] Independently for each $\BERT$ layer, we find a gender direction that encodes gender information.
 
 \item[2] We evaluate the quality of obtained directions by defining a new metric gender separability.
 
 \item[3] Subsequently, we propose Algorithm-\ref{Algorithm:singlepm} to obtain fine-grained gender directions and to introduce a new setting--$\BERT^{\text{De}}$--which lacks in gender-rich features.
 
\end{itemize}

Following \citet{bolukbasi2016man} work on identifying gender axis (direction) in context-independent word embedding, we extend it to extract geometric directions from contextualized word embeddings of $\BERT$. We hypothesize--\textsl{For every layer in BERT, there exists a low-dimensional context-independent subspace that encodes gender information.}

Thus, for each $\BERT$ layer $layer_k$ ($0\leq k \leq 12$), we aim to capture a d-dimensional gender subspace $B_k$ spanned by the basis vectors \{$b_k^1, \ldots, b_k^d$\} $\in {\rm I\!R}^{d_b}$. We define the basis-vectors as \textit{gender directions}. Ideally, the difference in vector representations of He and She should show a major component in gender directions. However, even in simple static embedding like Glove, such vectors may not behave as expected \cite{bolukbasi2016man}. The task of revealing gender subspace becomes even more difficult in case of contextualized embeddings, i.e., a word can map to more than one vector representations depending on the context. Such embeddings may lead to inconsistency in extracted directions, and thus, subspace. We propose a way to identify a static gender subspace $B_k$ by enforcing context to have many gender-specific words, all of which represent the same gender. We further elaborate on the method below.

\paragraph{Definition pair}
Let $O_g \coloneqq \{(f_i,m_i)\}_{i=1}^g$ be the ordered pair of words. The $f_i$ represents a noun that is commonly used for a female. Similarly, $m_i$ carries a male notion~\footnote{We focus on those words having low word-sense ambiguity}. Using $O_g$, we form a definition pair of sentences:

\begin{align*}
 S_f &= w_1 \ldots f_1 \ldots f_g \ldots w_n\ \\ S_m &=w_1 \ldots m_1 \ldots m_g \ldots w_n
\end{align*}

The definition pair makes use of the $O_g$ in a close context. We denote word at position $i$ in sequence $S_f$ and $S_m$ as $S_f^i$ and $S_m^i$. The definition set ($S_f^i$, $S_m^i$) satisfies either of the two conditions:
\begin{itemize}
    \item[1] ($S_f^i$, $S_m^i$) $\in$ $O_g$;
    \item[2] $S_f^i$ = $S_m^i$, if $S_f^i$ and $S_m^i$ are gender-neutral word.
\end{itemize}

From 10 gender pairs introduced in \citet{bolukbasi2016man}, we chose 9 and added \textsl\{Queen,King\} and \textsl\{Aunt,Uncle\}. Thus, $O_g$ contains 11 gender pairs. Our experimental findings suggest that more number of gender pairs make it difficult to find principal directions. As mentioned before, we use $O_g$ along with gender-neutral words to generate $S_f$ and $S_m$~\footnote{(see Appendix).}.

Let $u_k^i$ and $v_k^i$ denote vector mapping of words $S_f^i$ and $S_m^i$ at $\BERT$ $layer_k$, respectively. Since $S_f$ and $S_m$ are the same except for gender-specific words, we expect their word vectors to have a close contextual relationship. Thus, we conjecture that the difference vector $D_k^i = \{v_k^i - u_k^i\}$ shows a noticeable shift in gender directions by canceling out other encoded information such as context and word position. Later, we empirically show the importance of gender directions extracted using difference vectors.

\subsection{Gender subspace}
Independently for each $layer_k$, Principal Component Analysis (PCA) over difference vectors $D_k \coloneqq \{D_k^i\}_{i=1}^{n}$ helps uncover gender subspace $B_k$. $\texttt{PCA}(D_k)$ returns $n$-orthogonal directions $p^1_k, \ldots, p^n_k$ in decreasing order of the explained variance (EV). Each direction encodes a different notion that supplements the abstract concept of gender. Higher EV signifies more crucial directions that define gender subspace. However, such directions may not behave as expected and encode other information unrelated to gender. Initially, we base our analysis on two principal directions i.e. $p^1_k, p^2_k$ forming the 2-dimensional gender subspace. Subsequently, we show both the gender directions have a significant overlap in the encoded information. This gives an intuition of primarily using the first principal component that leads to a 1-dimensional subspace for gender.

\paragraph{Gender Separability}
Consider a direction $l$ and vector representation of words $g_1,g_2$ as $x$ and $y$, where $g_1$ is a masculine word and $g_2$ is feminine, or vice versa. We can find a real value $c$ such that:

\begin{equation}
 \langle x,l\rangle \geq c\,{\text{ and }}\langle y,l\rangle \leq c
\end{equation}

$c$ divides $l$ in two rays (half spaces) each of which represents a unique gender. An ideal gender direction should project a word-vector in gender-specific ray. Thus, we define gender-separability as the accuracy of projection-based classification on the corresponding ray.

\paragraph{Gender Classification Dataset (Gen-data)} We compiled train-set from \cite{zhao2018learning}\footnote{\url{https://github.com/uclanlp/gn_glove/tree/master/wordlist}}. The dataset consists of 222 gender-word pairs, i.e., for each feminine word, there is a masculine counterpart. To form test-set, we collect gender-specific words from another source containing 595 male-specific and 404 female-specific words~\footnote{\url{https://github.com/ecmonsen/gendered_words}}. From this source, we collected 595 neutral words and randomly split them to assign 222 samples for training and rest for testing. \cref{tab:dataset_gender} shows Gen-data statistics and samples.

\begin{table}[t]
 \small
 \center
 \begin{tabular}{cccc}
 \toprule
  Data & Gender & \#Samples &Examples\\
 \midrule
 \multirow{6}{*}{Train} & \multirow{2}{*}{Female} & \multirow{2}{*}{222} & actress,mama,madam,\\
 &&&princess,sororal \\
 
 & \multirow{2}{*}{Male} & \multirow{2}{*}{222} & actor,papa,sir,\\
 &&&prince,fraternal \\
 
 & \multirow{2}{*}{Neutral} & \multirow{2}{*}{222} & guest, beast, friend\\
 &&&mentor, outlier \\

 \multirow{6}{*}{Test} & \multirow{2}{*}{Female} & \multirow{2}{*}{404} & chatelaine, ballerina, baroness,\\
&&&barmaid, brunette\\

 & \multirow{2}{*}{Male} & \multirow{2}{*}{595} & adonis, barman,baron,\\
 &&&brunet, charon \\
  
 & \multirow{2}{*}{Neutral} & \multirow{2}{*}{5701} & abator, owner, bidder,\\
 &&&genius, whistler \\
 
 \bottomrule
 \end{tabular}
 \caption{\footnotesize{Distribution of Gen-data.}}
 \label{tab:dataset_gender}
\end{table}

\begin{figure}[t]
 \centering
 \includegraphics[width=0.35\textwidth]{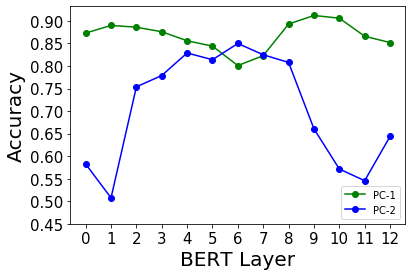}
 \caption{\footnotesize{Layer-wise gender separability on Gen-data when word vectors are projected on first principal component (PC-1) and second principal component (PC-2).}}
 \label{fig:PC_acc}
\end{figure}

\begin{figure}[t]
 \centering
 \includegraphics[width=0.35\textwidth]{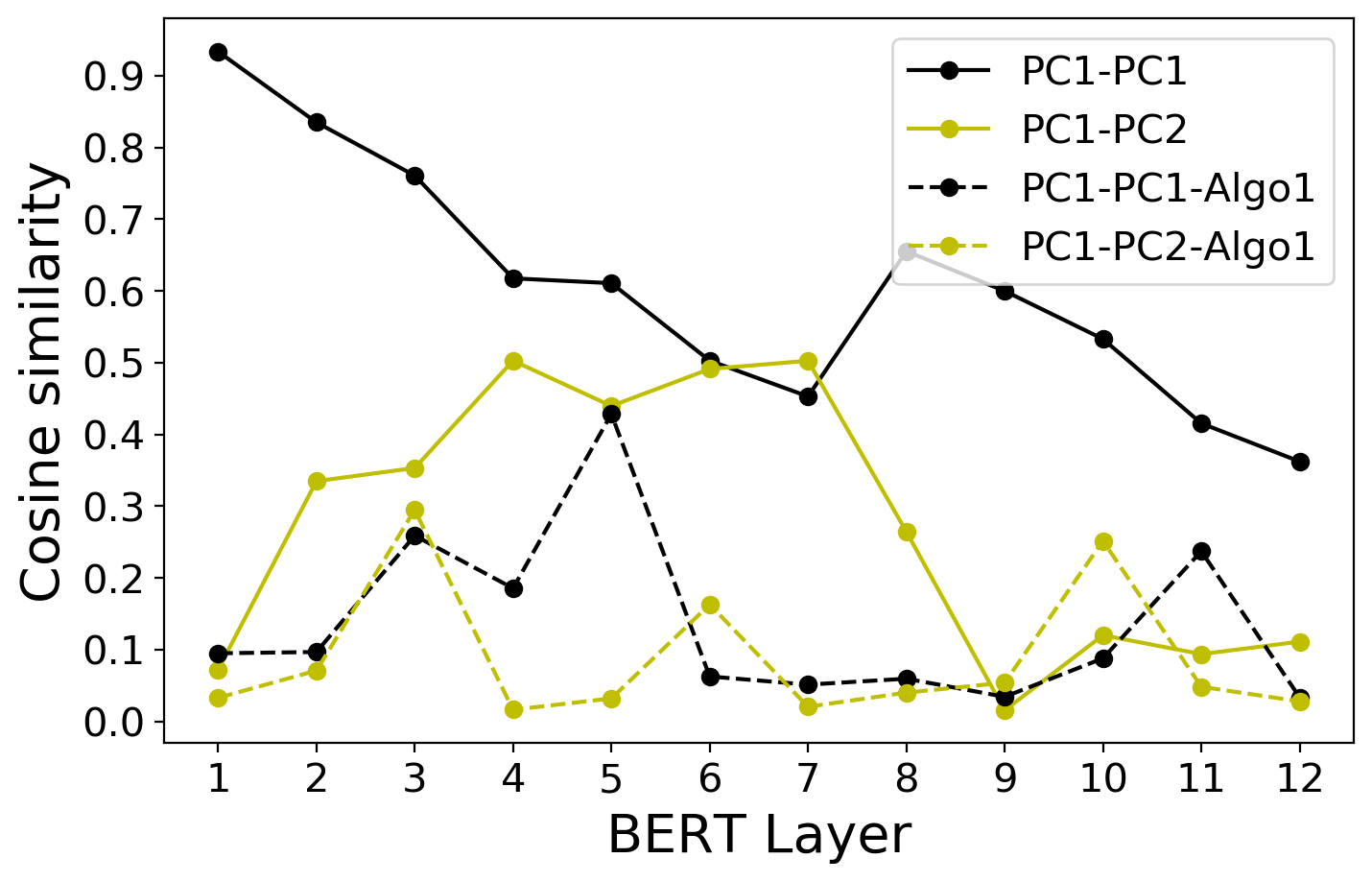}
 \caption{Cosine similarity between two vectors $Cos(u,v) = \frac{u \cdot v}{||u||\;||v||}$. PC1-PC1, PC1-PC-2 denotes $Cos$($p^1_0$, $p^1_j$) and $Cos$($p^1_0$, $p^2_j$), respectively. Algo1 denotes PCs obtained using Algorithm\ref{Algorithm:singlepm}.}
 \label{fig:PCs}
\end{figure}

We evaluate layer-specific gender subspaces for their gender separability on Gen-Data, we evaluate the first two principal directions $p^1_k$ and $p^2_k$ of a given $layer_k$. To find layer-specific $c$, i.e., $c_k^i$ for $i_{th}$ principal component (PC), we perform a grid search to maximize separability on the train-set of Gen-data excluding gender-neutral words. We then test the quality of separation on its test-set as shown in the \cref{fig:PC_acc}. We observe the first PC of all layers have high separability score. Moreover, second PCs of middle layers performs as good as respective first PCs. This observation raises a question:- is second gender direction, i.e., $p^2_k$ crucial to define the gender subspace? We answer it by the following analysis: 

Since all the contextual embeddings are transformations of embedding at $layer_0$, in \cref{fig:PCs}, we plot the cosine similarity between first principal gender direction of $layer_0$ ($p^1_0$) and $layer_j$ ($p^1_j$), where $1\leq j \leq 12$. The high cosine similarity~\footnote{cosine of the angle between two random vectors in high dimensions is zero with high probability.} depicts passing of encoded gender information from $layer_0$ and lesser new gender-specific features learned by following layers. We also analyse cosine similarity between $p^1_0$ and $p^2_j$ which is the second principal gender direction at $layer_j$. The similarity score increases in the middle layers which supports our observation in \cref{fig:PC_acc} with high separability values. Moreover, it also indicates that $p^2_j$'s hardly encode any extra gender-specific information keeping aside what is acquired from $p^1_0$.
 
Eliminating vector components in gender directions is expected to reduce gender-bias in downstream tasks. However, due to non-ideal behavior, omitting a large number of directional components may cause representation noise and hinder the quality of $\BERT$ embeddings. Moving forward, we propose Algorithm-\ref{Algorithm:singlepm} that aims to tackle this issue.

\begin{algorithm}

\small
\SetKwFunction{cumprod}{cumprod}
\SetKwFunction{length}{length}
\SetKwFunction{zeros}{zeros}
\SetKwFunction{ceil}{ceil}

\SetKwInOut{Input}{Input}
\SetKwInOut{Output}{Output}

\caption{{Extracting layer-wise principal component in Gender subspace.}\label{Algorithm:singlepm}}

\Input{
        \xvbox{3mm}-Strings pair ($S_m$, $S_f$), which differ only in gender-specific words.\\
     }
\Output{
        \xvbox{3mm}-{$\xvar{\textit{P}}$}=Layer-wise principal component set $\{P_0, \ldots, P_{12}\}$.
      }

 \BlankLine

 \xvbox{6.0mm}{$W_{tf}$} $\leftarrow$ \texttt{Tokenize}($S_f$) \tcc*{WP Tokenization}

 \xvbox{6.0mm}{$W_{tm}$} $\leftarrow$ \texttt{Tokenize}($S_m$) \tcc*{WP Tokenization}

 \xvbox{3.0mm}{$u_0$} $\leftarrow$ \texttt{Layer}$_0$($W_{tf}$) \tcc*{Context-independent input vectors for $S_f$}

 \xvbox{3.0mm}{$v_0$} $\leftarrow$ \texttt{Layer}$_0$($W_{tm}$) \tcc*{Context-independent input vectors for $S_m$}

 \xvbox{3.0mm}{$D_0$} $\leftarrow$ ($v_0-u_0$) \tcc*{Difference vector}

 \xvbox{3.0mm}{$P_0$} $\leftarrow$ \texttt{PCA}($D_0$) \tcc*{PC with maximum EV}

 \For{$\xvar{j}\leftarrow$ $[ 1,2,\dots, 12 ]$ }{

 \xvbox{3.0mm}{$u_{j-1}^*$} $\;\;\;\;\leftarrow$ \texttt{Proj}$_{\perp P_{j-1}}(u_{j-1})$ \tcc*{Perpendicular projection}
 
 \xvbox{3.0mm}{$v_{j-1}^*$} $\;\;\;\;\leftarrow$ \texttt{Proj}$_{\perp P_{j-1}}(v_{j-1})$
 
 \xvbox{3mm}{$u_j$} $\leftarrow$ \texttt{Layer}$_\text{j}$($u_{j-1}^*$)
 
 \xvbox{3mm}{$v_j$} $\leftarrow$ \texttt{Layer}$_\text{j}$($v_{j-1}^*$)
 
 \xvbox{3.0mm}{$D_j$} $\leftarrow$ ($v_j-u_j$) \tcc*{Difference vector}

 \xvbox{3.0mm}{$P_j$} $\leftarrow$ \texttt{PCA}($D_j$)
 } 

\end{algorithm}

\subsection{Reducing Gender Bias}
For each layer in $\BERT$, Algorithm-\ref{Algorithm:singlepm} extracts first principal gender directions $p_j^1$ (referred as $P_j$) in a systematic way. The $layer_0$ maps WordPiece tokens of $S_f$ and $S_m$ to a set of vectors $u_i$ and $v_i$, respectively. PCA over difference vector $D_0 = (v_0-u_0)$ gives $P_0$ i.e. gender direction with maximum explained variance. We remove components of $v_0$ and $u_0$ on $P_0$ by taking perpendicular projections. For a vector $a$, the projection perpendicular to a unit vector $b$ is defined as:

\begin{equation}
    Proj_{\perp b}(a) \coloneqq a - \langle a, b\rangle b
\end{equation}

\begin{center}
    (Where $\langle a,b \rangle$ is inner product of vectors a and b.)
\end{center}

We feed the projected vectors $v_0^{*}$ and $u_0^{*}$ to the next layer, i.e., $layer_1$. The same procedure is repeated until final layer $layer_{12}$ and all the extracted principal components $P \coloneqq {P_1, \ldots, P_{12}}$ are stored. It is worthwhile pointing that the algorithm is different from independent layer-wise analysis as each layer has missing gender information from layers preceding it. The new cosine similarity scores show a significant drop in \cref{fig:PCs} - dotted.

\begin{table*}[t]
 \centering
 \resizebox{\textwidth}{!}{
 \begin{tabular}{|c|ccc|ccc|ccc|ccc|}
 \hline
 \multirow{3}{*}{Emotion} & \multicolumn{6}{c|}{Emotion Intensity} & \multicolumn{6}{c|}{Valence Intensity}  \\ 
 \cline{2-13}&\multicolumn{3}{c|}{$\BERT$} &\multicolumn{3}{c|}{$\BERT^{\text{De}}$}&\multicolumn{3}{c|}{$\BERT$} &\multicolumn{3}{c|}{$\BERT^{\text{De}}$}\\
 \cline{2-13}
&Pearson &$\Delta_{F{\uparrow}{-}M{\downarrow}}$ &$\Delta_{M{\uparrow}{-}F{\downarrow}}$ &Pearson &$\Delta_{F{\uparrow}{-}M{\downarrow}}$(\%d) &$\Delta_{M{\uparrow}{-}F{\downarrow}}$(\%d)&Pearson &$\Delta_{F{\uparrow}{-}M{\downarrow}}$ &$\Delta_{M{\uparrow}{-}F{\downarrow}}$ &Pearson &$\Delta_{F{\uparrow}{-}M{\downarrow}}$(\%d) &$\Delta_{M{\uparrow}{-}F{\downarrow}}$(\%d)\\
\hline
 Joy &0.666 &0.0396 &0.0402 &0.660 &\;0.0143(${\downarrow}$\textbf{63.9})\;&\;0.0143(${\downarrow}$\textbf{64.4})\;& \multirow{4}{*}{0.659} & 0.0346 &0.0376 & \multirow{4}{*}{0.670} & \;0.0209(${\downarrow}$\textbf{39.5}) \;&\;0.0138(${\downarrow}$\textbf{63.3})\;\\
 Fear &0.581 &0.0202 &0.0244 &0.593 &\;0.0152(${\downarrow}$\textbf{24.7})\;&\;0.0158(${\downarrow}$\textbf{35.2})\;&&0.0263 &0.0244 &&\;0.0156(${\downarrow}$\textbf{40.6}) \;&\;0.0123(${\downarrow}$\textbf{49.5})\;\\
 Sadness &0.615 &0.0380 &0.0138 &0.604 &\;0.0178(${\downarrow}$\textbf{58.9})\;&\;0.0097(${\downarrow}$\textbf{29.7})\;&&0.0272 &0.0205 &&\;0.0153(${\downarrow}$\textbf{43.7}) \;&\;0.0118(${\downarrow}$\textbf{42.4})\;\\
 Anger &0.627 &0.0074 &0.0316 &0.626 &\;0.0121(${\uparrow}63.5$)\;&\;0.0149(${\downarrow}$\textbf{52.8})\;&&0.0219 &0.0198 &&\;0.0130(${\downarrow}$\textbf{40.6}) \;&\;0.0119(${\downarrow}$\textbf{39.8})\;\\
 \hline
 \end{tabular}
 }
 \caption{Final-layer ($layer_{12}$) of $\BERT$ and $\BERT^{\text{De}}$ equity evaluation of the five-intensity regression models. \%d refers to the percentage change in $\Delta$ values. The p-values for 1) Emotion intensity models: \{anger\}${\leq0.05} \; ({\geq}{0.70^*})$; \{joy, fear, sad\} $\leq{0.20} \; ({\geq}{0.70^*})$. 2) The valence intensity model (emotion-wise p-values): \{anger\}${\leq0.05} \; ({\geq}{0.75^*})$; \{joy, fear, sad\}$\leq{0.20}\;({\geq}{0.70^*})$, where values with * denotes $\BERT^{\text{De}}$-based $\MLP$ regressor.}
 
 \label{tab:EEC_Biased_Emt}
\end{table*}

\begin{table*}[t]
 \centering
 \resizebox{\textwidth}{!}{
 \begin{tabular}{|c|cccc|cccc|cccc|cccc|}
 \hline
 \multirow{3}{*}{Emotion}& \multicolumn{8}{c|}{Emotion Intensity} & \multicolumn{8}{c|}{Valence Intensity} \\
 \cline{2-17}&\multicolumn{4}{c|}{$\BERT$} &\multicolumn{4}{c|}{$\BERT^{\text{De}}$}&\multicolumn{4}{c|}{$\BERT$} &\multicolumn{4}{c|}{$\BERT^{\text{De}}$}\\
 \cline{2-17}
&$\#_{F{\uparrow}{-}M{\downarrow}}$ &$\#_{M{\uparrow}{-}F{\downarrow}}$ 
&$\delta$ 
&($\#_{F{=}M}$) 
&$\#_{F{\uparrow}{-}M{\downarrow}}$ &$\#_{M{\uparrow}{-}F{\downarrow}}$
&$\delta$
&($\#_{F{=}M}$)
&$\#_{F{\uparrow}{-}M{\downarrow}}$ &$\#_{M{\uparrow}{-}F{\downarrow}}$ 
&$\delta$ 
&($\#_{F{=}M}$) 
&$\#_{F{\uparrow}{-}M{\downarrow}}$ &$\#_{M{\uparrow}{-}F{\downarrow}}$
&$\delta$
&($\#_{F{=}M}$)
\\
\hline
 Joy &92 &291 &198 &2 &197 &177 &20(${\downarrow}$\textbf{178}) &11(${\uparrow}$\textbf{9})&62 &322 &260 &1 &227 &147 &80(${\downarrow}$\textbf{180}) &11(${\uparrow}$\textbf{11})\\
 Fear &177 &204 &27 &4 &175 &199 &24(${\downarrow}$\textbf{3}) &10(${\uparrow}$\textbf{6})&105 &276 &171 &4 &207 &165 &42(${\downarrow}$\textbf{129}) &13(${\uparrow}$\textbf{9})\\
 Sadness &339 &44 &294 &2 &243 &128 &114(${\downarrow}$\textbf{180}) &14(${\uparrow}$\textbf{12})&106 &274 &168 &5 &209 &162 &47(${\downarrow}$\textbf{121}) &14(${\uparrow}$\textbf{9})\\
 Anger &18 &366 &347 &1 &161 &212 &52(${\downarrow}$\textbf{295}) &12(${\uparrow}$\textbf{11})&126 &258 &132 &1 &229 &148 &81(${\downarrow}$\textbf{51}) &8(${\uparrow}$\textbf{7})\\
 \hline
 \end{tabular}
 }
 \caption{Final layer of $\BERT$ and $\BERT^{\text{De}}$ equity evaluation: $F{\uparrow}{-}M{\downarrow}$, $M{\uparrow}{-}F{\downarrow}$, and $M{=}F$. $\delta$ = $|\#_{F{\uparrow}{-}M{\downarrow}}- \#_{M{\uparrow}{-}F{\downarrow}}|$}
 \label{tab:EEC_counts_Emt}
\end{table*}

\begin{table*}[ht!]
 \centering
 \resizebox{\textwidth}{!}{
 \begin{tabular}{|c|ccc|ccc|ccc|ccc|}
 \hline
 \multirow{3}{*}{Emotion} & \multicolumn{6}{c|}{Emotion Intensity} & \multicolumn{6}{c|}{Valence Intensity}  \\ 
 \cline{2-13}&\multicolumn{3}{c|}{$\BERT$} &\multicolumn{3}{c|}{$\BERT^{\text{De}}$}&\multicolumn{3}{c|}{$\BERT$} &\multicolumn{3}{c|}{$\BERT^{\text{De}}$}\\
 \cline{2-13}
&Pearson &$\Delta_{F{\uparrow}{-}M{\downarrow}}$ &$\Delta_{M{\uparrow}{-}F{\downarrow}}$ &Pearson &$\Delta_{F{\uparrow}{-}M{\downarrow}}$(\%d) &$\Delta_{M{\uparrow}{-}F{\downarrow}}$(\%d)&Pearson &$\Delta_{F{\uparrow}{-}M{\downarrow}}$ &$\Delta_{M{\uparrow}{-}F{\downarrow}}$ &Pearson &$\Delta_{F{\uparrow}{-}M{\downarrow}}$(\%d) &$\Delta_{M{\uparrow}{-}F{\downarrow}}$(\%d)\\
\hline
 Joy &0.580 &0.0436 &0.0152 &0.557 &0.0195\;(${\downarrow}$\textbf{55.2})\;&\;0.0165(${\uparrow} 8.75$)\;& \multirow{4}{*}{0.658} & 0.0356 &0.0118 & \multirow{4}{*}{0.653} & \;0.0118(${\downarrow}$\textbf{66.6}) \;&\;0.0086(${\downarrow}$\textbf{26.65})\;\\
 Fear &0.475 &0.0256 &0.0241 &0.497 &\;0.0139(${\downarrow}$\textbf{45.4})\;&\;0.0130(${\downarrow}$\textbf{ 45.6})\;&&0.0348 &0.0113 &&\;0.0117(${\downarrow}$\textbf{66.2}) \;&\;0.0099(${\downarrow}$\textbf{11.8})\;\\
 Sadness &0.532 &0.0282 &0.0129 &0.535 &\;0.0156(${\downarrow}$\textbf{44.4})\;&\;0.0133(${\uparrow} 2.8$)\;&&0.0192 &0.0089 &&\;0.0185(${\downarrow}$\textbf{3.38}) \;&\;0.0113(${\uparrow}26.6$)\;\\
 Anger &0.571 &0.0123 &0.0408 &0.577 &\;0.0133(${\uparrow}8.6$)\;&\;0.0124(${\downarrow}$ \textbf{69.4})\;&&0.0185 &0.0109 &&\;0.0177(${\downarrow}$\textbf{4.24}) \;&\;0.0115(${\uparrow}5.57$)\;\\
 \hline
 \end{tabular}
 }
 \caption{Pre-final layer ($layer_{11})$ of $\BERT$ and $\BERT^{\text{De}}$ equity evaluation of the five-intensity regression models. \%d refers to the percentage change in $\Delta$ values. The p-values for 1) Emotion intensity models: \{fear\}${\leq0.01} \; ({\geq}{0.95^*})$; \{anger, joy, sad\} $\leq{0.20} \; ({\geq}{0.85^*})$. 2) The valence intensity model (emotion-wise p-values): \{anger, fear\}${\leq0.05} \; ({\geq}{0.95^*})$; \{joy, sad\}$\leq{0.20}\;({\geq}{0.85^*})$.}
 \label{tab:EEC_Biased_Emt_pref}
\end{table*}

\begin{table*}[ht!]
 \centering
 \resizebox{\textwidth}{!}{
 \begin{tabular}{|c|cccc|cccc|cccc|cccc|}
 \hline
 \multirow{3}{*}{Emotion}& \multicolumn{8}{c|}{Emotion Intensity} & \multicolumn{8}{c|}{Valence Intensity} \\
 \cline{2-17}&\multicolumn{4}{c|}{$\BERT$} &\multicolumn{4}{c|}{$\BERT^{\text{De}}$}&\multicolumn{4}{c|}{$\BERT$} &\multicolumn{4}{c|}{$\BERT^{\text{De}}$}\\
 \cline{2-17}
&$\#_{F{\uparrow}{-}M{\downarrow}}$ &$\#_{M{\uparrow}{-}F{\downarrow}}$ 
&$\delta$ 
&($\#_{F{=}M}$) 
&$\#_{F{\uparrow}{-}M{\downarrow}}$ &$\#_{M{\uparrow}{-}F{\downarrow}}$
&$\delta$
&($\#_{F{=}M}$)
&$\#_{F{\uparrow}{-}M{\downarrow}}$ &$\#_{M{\uparrow}{-}F{\downarrow}}$ 
&$\delta$ 
&($\#_{F{=}M}$) 
&$\#_{F{\uparrow}{-}M{\downarrow}}$ &$\#_{M{\uparrow}{-}F{\downarrow}}$
&$\delta$
&($\#_{F{=}M}$)
\\
\hline
 Joy &310 &74 &236 &1 &199 &177 &22(${\downarrow}$\textbf{214}) &9(${\uparrow}$\textbf{8})&328 &55 &273 &2 &188 &186 &2(${\downarrow}$\textbf{271}) &11(${\uparrow}$\textbf{9})\\
 Fear &223 &160 &63 &2 &159 &212 &53(${\downarrow}$\textbf{10}) &14(${\uparrow}$\textbf{12}) &335 &45 &290 &5 &194 &172 &22(${\downarrow}$\textbf{268}) &19(${\uparrow}$\textbf{14})\\
 Sadness &281 &99 &182 &5 &210 &169 &41(${\downarrow}$\textbf{141}) &6(${\uparrow}$\textbf{1})&283 &85 &198 &17 &202 &173 &29(${\downarrow}$\textbf{169}) &10(${\uparrow}$\textbf{7})\\
 Anger &31 &352 &321 &2 &145 &225 &80(${\downarrow}$\textbf{241}) &15(${\uparrow}$\textbf{13})&269 &109 &160 &7 &217 &155 &62(${\downarrow}$\textbf{98}) &13(${\uparrow}$\textbf{6})\\
 \hline
 \end{tabular}
 }
 \caption{Pre-final layer $\BERT$ and $\BERT^{\text{De}}$ equity evaluation: Number of occurrences of $F{\uparrow}{-}M{\downarrow}$, $M{\uparrow}{-}F{\downarrow}$, and $M{=}F$. $\delta$ = $|\#_{F{\uparrow}{-}M{\downarrow}}- \#_{M{\uparrow}{-}F{\downarrow}}|$.}
 \label{tab:EEC_counts_Emt_pref}
\end{table*}

\paragraph{Removing Gender Component}
After obtaining layer-wise gender directions $P$, we introduce a new $\BERT$ setting--$\BERT^{\text{De}}$. As an enhancement of $\BERT$, $\BERT^{\text{De}}$ removes gender components from a token's vector representations. Given an input sequence of tokens $W_t$ to the $layer_0$, we obtain token-vectors $W_0$ at its output. For each vector $t_0^i$ in  $W_0$, we remove its component in direction $P_0$, i.e., $t_0^{i*} = Proj_{\perp P_0}(t_0^i)$. We denote the set of $t_0^{i*}$ vectors as $W_0^*$. Unlike normal $\BERT$ settings which feed $W_0$ as input to $Layer_1$, we feed $W_0^*$. We iterate this process for every layer $layer_j$ ($0{\leq}j{\leq}12$) which receives $W_{j-1}^*$ ($j>0$) at input and gives $W_j^*$ at output by removing its vector components in direction $P_j$. In the next section, we evaluate $\BERT^{\text{De}}$ on EEC and compare its performance with $\BERT$.

\section{Equity Evaluation of $\text{BERT}^{\text{De}}$ - MLP}
We follow the same methodology as in \cref{BERT-MLP}, however, by substituting $\BERT$ with $\BERT^{\text{De}}$. For each task-specific $\MLP$ trained on $\BERT^{\text{De}}$ $layer_{11}$ and $layer_{12}$, we perform paired two-sample t-tests to determine whether the mean difference between male and female scores is significant. Low p-values indicate a significant difference in model predictions based on gender. 

\subsection{Results and Discussion}
As shown in the \cref{tab:EEC_Biased_Emt} and \ref{tab:EEC_Biased_Emt_pref}, most of the $\BERT^{\text{De}}$-$\MLP$ regression models show an overall \% decrease in $\Delta$ values in both $F{>}M$ ($\Delta_{F{\uparrow}{-}M{\downarrow}}$) and $M{>}F$ ($\Delta_{M{\uparrow}{-}F{\downarrow}}$) cases. Final-layer $\BERT$-$\MLP$ models for joy, fear, and anger have higher average intensity scores for male phrases than female, while the opposite trend is seen in models for valence and sadness. It is also noteworthy that the pre-final layer ($layer_{11}$) shows a somewhat opposite trend. Hence, we suspect the $\BERT$-induced bias depends on the which layer embedding is used. Moreover, from p-values of $\BERT$-based regressors, we see much higher significant $\Delta$ values as compared to regressors using $\BERT^{\text{De}}$. From \cref{tab:EEC_counts_Emt} and \ref{tab:EEC_counts_Emt_pref}, for all five regressors, we observe a significant reduction in difference between number occurrences where $F{>}M$ and $M{>}F$, i.e., $\delta$. We also observe an increase in cases when regressors assign equal scores to both genders, i.e., $\#F{=}M$. Unlike $\BERT$, models based on $\BERT^{\text{De}}$ show no consistency in assigning higher intensity scores to either male or female. Hence, simple $\MLP$ regressors based on $\BERT^{\text{De}}$ vectors show an apparent gender unbiased nature on EEC.

\subsection{Semantic Consistency}
Gender debiasing of a model is desirable, however, it may come at the cost of reduced model performance on the task. In the case of $\BERT^{\text{De}}$, removal of component in identified directions can lead to a loss of other semantic information. Thus, to check the semantic consistency of $\BERT^{\text{De}}$, we compared Pearson's correlation score of $\BERT$-$\MLP$ and $\BERT^{\text{De}}$-$\MLP$ regressors on the task-specific test-sets. As depicted in the \cref{tab:EEC_Biased_Emt} and \ref{tab:EEC_Biased_Emt_pref}, there is no drastic reduction in Pearson's scores, confirming semantic is preserved. Thus, removing the directional components reduces the gender-bias induced by $\BERT$, while maintaining the regressors performance on the downstream tasks. Next, we define a gender classification task to investigate the relevance of extracted directions, i.e., how informative they are about the gender of a given word.

\section{Evaluation on Gender Classification}
It is evident from the above analysis that $\BERT^{\text{De}}$ is effective in reducing gender-biased predictions of $\MLP$s. Moreover, the semantic consistency proves $\BERT^{\text{De}}$ to be as effective as $\BERT$ on all the five tasks. To substantiate that Algorithm-\ref{Algorithm:singlepm} makes $\BERT^{\text{De}}$ word embedding deficient in gender-specific features, we design another downstream task - gender classification of a word. A naive solution to reduce gender-bias is to remove all gender-specific words. We analyse the suitability of this solution in the end.

\subsection{Baseline Gender Classifier (GC)}
First, we establish a gender classification baseline to compare the performances of $\BERT$ and $\BERT^{\text{De}}$. Input to the baseline is WordPience tokenized ${w}{~}{\rightarrow}~\{w_1,\ldots,w_N\}$. We randomly initialize the WordPiece embeddings - $E{:}\;{w_i}\:{\mapsto}\;{t_i}\:{\in}\:{{\rm I\!R}^{100}}{\sim}{\mathcal{U}(-1,1)}$. The possibility of multiple subwords makes it intuitive to perform convolution over $t_i$'s \cite{kim-2014-convolutional}. The 1-dimensional convolution layer \textsl{Conv-1D} consists of 32 filters each of size 1. Thus the input tensor of shape $N \times 100$ after convolution at stride 1 transforms to $N \times 32$; this followed by global max-pooling gives $1 \times 32$ feature vector. The vector is passed through a fully connected layer consists of 128 neurons, and an output layer with $sigmoid$ activation. We minimize the categorical cross-entropy of the output against the target gender set $\mathcal{L} \in \{0,1\}^{|N|}$, where $N$: number of data samples, 1 and 0 are input labels for female and male, respectively (2-way classification). We use Adam optimizer with learning rate 0.001. We randomly drop-out $20\%$ of \textsl{FC} layer activations to prevent parameter overfitting. Hyperparameters are tuned to maximize average 10-fold cross-validation accuracy. 

\begin{figure}
 \centering
 \includegraphics[width=0.4\textwidth]{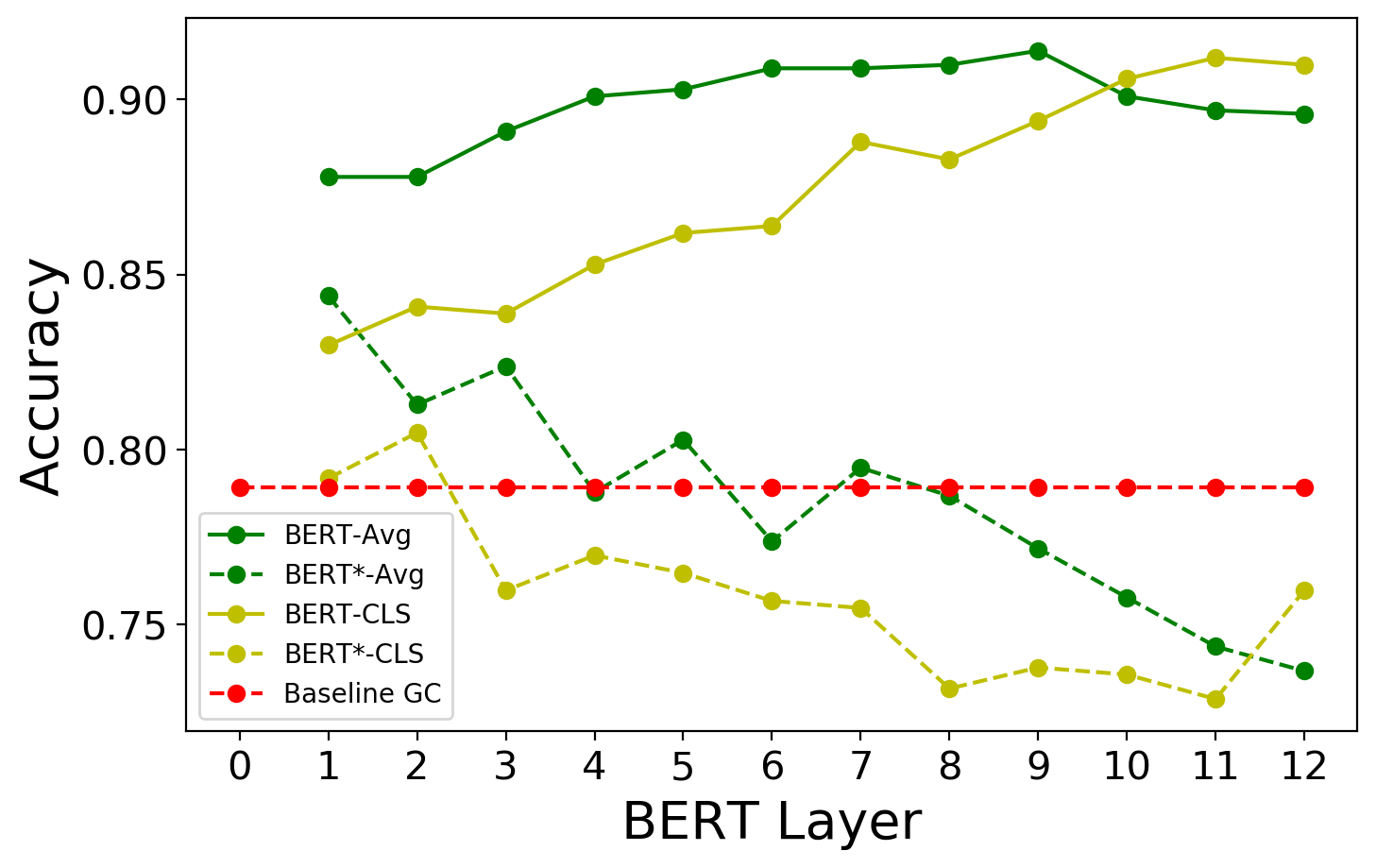}
 \caption{\footnotesize{$\BERT$-CLS and $\BERT^{\text{De}}$-CLS denote $\MLP$ accuracy using $layer_k$ (x-axis) vectors in $I_1$ setting. Similarly, $\BERT$-Avg and $\BERT^{\text{De}}$-Avg refer to the $I_2$ setting. Switching from $\BERT$ to $\BERT^{\text{De}}$, we see a significant drop in $\MLP$ gender-classification performance in both $I_1$ and $I_2$ inputs cases.}}
 \label{fig:probe_results_debias}
\end{figure}

\subsection{MLP Gender Classifier}
Similar to~\citet{tenney2019you}, we create a 2-layer $\MLP$ classifier. The architecture is very similar to \cref{fig:mlp_regressor} except for $\MLP$ is used for classification. The input to $\BERT$ is a word $w$. Given the output of $\BERT$ $layer_k$, i.e., $(t^{cls}_k, t^1_k, \ldots, t^n_k, t^{sep}_k)$, we analyse two different input settings to $\MLP$:

\begin{itemize}
\setlength\itemsep{0.01em}
\item $I_1:$ Vector representation of \texttt{[CLS]}, i.e., $t_k^{cls}$.
\item $I_2:$ Average of all token vectors ${t_k^{Avg}\:}{\coloneqq}{\frac{t_k^{cls}+\sum_{i=1}^{n} t_k^i +t_k^{sep}}{(n+2)}}$.
\end{itemize}

For a $layer_k$ and input setting, we train a separate $\MLP$ on the train-set of Gen-data and evaluate on its test-set. Each $\MLP$ takes $I_1$ or $I_2$ at input and predicts gender. Thus, given $\BERT$ model, we train-test 24 $\MLP$s ($12{\times}2$). The $\MLP$s have 100 hidden layer neurons. We determine hyperparameters using a validation set comprised of the 10\% samples from the training set. Rest settings are similar to the baseline.

\subsection{BERT-MLP vs $\text{BERT}^{\text{De}}$ - MLP}
Following the above-mentioned method, we evaluate $\MLP$ classifiers based on $\BERT$ and $\BERT^{\text{De}}$ embeddings. As shown in the \cref{fig:probe_results_debias}, we find $\BERT$-based $\MLP$s outperforming the gender classification baseline in both the setting $I_1$ and $I_2$. This depicts the existence of gender-rich features in $\BERT$ provided embeddings. $\BERT^{\text{De}}$-$\MLP$ shows much poorer performance as compared to $\BERT$-$\MLP$. This observation makes it clear that removed directional components from the embeddings omit gender-rich features, hence, the obtained directions have a high magnitude of cosine similarity with actual gender directions.
Moreover, $\BERT^{\text{De}}$-MLP accuracy drops even below baseline at deeper layers, suggesting the similarity magnitude increases as the embeddings become deeply contextualized.

\begin{table}[ht!]
 \centering
\small
 \begin{tabular}{|c| c| c| c| c|}
\hline
\multirow{2}{*}{Layer} &\multicolumn{2}{c|}{$\BERT$} &\multicolumn{2}{c|}{$\BERT^{\text{De}}$}\\
\cline{2-5}
&2-way &3-way &2-way &3-way \\
\hline
0 &81.4 &83.7 &80.4(${\downarrow}1.0$) &81.4(${\downarrow}2.3$)\\
1 &83.3 &85.9 &81.4(${\downarrow}1.9$) &81.6(${\downarrow}4.3$)\\
2 &80.4 &85.1 &75.9(${\downarrow}4.5$) &79.0(${\downarrow}6.1$)\\
3 &79.8 &84.3 &74.9(${\downarrow}4.9$) &83.1(${\downarrow}1.2$)\\
4 &82.4 &85.4 &75.8(${\downarrow}7.1$) &82.3(${\downarrow}3.2$)\\
5 &82.9 &85.5 &75.8(${\downarrow}7.1$) &81.2(${\downarrow}4.3$)\\
6 &86.8 &86.8 &74.8(${\downarrow}12.0$) &81.2(${\downarrow}5.6$)\\
7 &86.4 &86.0 &81.2(${\downarrow}5.2$) &79.1(${\downarrow}6.9$)\\
8 &86.5 &89.1 &82.8(${\downarrow}3.7$) &70.9(${\downarrow}18.2$)\\
9 &84.3 &87.3 &79.5(${\downarrow}4.8$) &71.9(${\downarrow}15.4$)\\
10 &81.6 &85.6 &72.8(${\downarrow}8.8$) &74.7(${\downarrow}10.9$)\\
11 &84.5 &87.1 &71.9(${\downarrow}12.6$) &74.8(${\downarrow}12.3$)\\
12 &85.4 &86.8 &71.9(${\downarrow}13.5$) &67.9(${\downarrow}18.9$)\\
\hline
GC &\multicolumn{2}{c|}{Male: 51.1} &\multicolumn{2}{c|}{Male: 50.3}\\
(Random) &\multicolumn{2}{c|}{Female: 48.9} &\multicolumn{2}{c|}{Female: 49.7} \\
\hline
 \end{tabular}
 \caption{\footnotesize{Percentage of misclassified neutral words predicted as Male.}}
 \label{tab:HDB}
\end{table}

Additionally, we train $\MLP$s on a 3-way classification task that includes gender-neutral words from Gen-data as a part of training the $\MLP$s and an extra category apart from female and male, i.e., neutral. \cref{tab:HDB} shows the percentage of neutral words misclassified in male class ($I_1$ - setting). It signifies that even for a gender-neutral word, $\BERT$ embeddings contain gender notion. Hence, simply removing the gender-specific words from the input sequence would not be a robust solution to tackle gender-bias in downstream applications. However, the misclassification percentage decreases in case of $\BERT^{\text{De}}$. Our proposed method does not need to avail any gender-specific information of an input word.

\section{Conclusion}
We studied gender-bias induced by $\BERT$ in five downstream tasks. Using PCA, we identified orthogonal directions -- defining a subspace -- in $\BERT$ word embeddings that encode gender informative features. We then introduced an algorithm to identify fine-grained gender directions, i.e., 1-dimensional gender subspace. Omitting word vector components in such directions proved to be reducing gender-bias in the downstream tasks. The method can be adapted to study other social biases such as race and ethnicity.

\bibliography{arxiv.bib}
\end{document}